# Transformer Models for Acute Brain Dysfunction Prediction

**Brandon Silva** [1,3*]**, Miguel Contreras** [1,3*]**, Tezcan Ozrazgat Baslanti** [2,3]**, Yuanfang Ren** [2,3]**, Ziyuan Guan**[2,3]**, Kia Khezeli** [1,3]**, Azra Bihorac** [2,3]**, Parisa Rashidi** [1,3**]

[1]Department of Biomedical Engineering, University of Florida, Gainesville, FL USA

[2]Department of Medicine, University of Florida, Gainesville, FL USA

[3]Intelligent Critical Care Center (IC3), University of Florida, Gainesville, FL USA

**\* Denotes equal contribution**

**\*\* Correspondence:**
Parisa Rashidi
parisa.rashidi@bme.ufl.edu

## Abstract

Acute brain dysfunctions (ABD), which include coma and delirium, are prevalent in the ICU, especially among older patients. The current approach in manual assessment of ABD by care providers may be sporadic and subjective. Hence, there exists a need for a data-driven robust system automating the assessment and prediction of ABD. In this work, we develop a machine learning system for real-time prediction of ADB using Electronic Health Record (EHR) data. Our data processing pipeline enables integration of static and temporal data, and extraction of features relevant to ABD. We train several state-of-the-art transformer models and baseline machine learning models including CatBoost and XGB on the data that was collected from patients admitted to the ICU at UF Shands Hospital. We demonstrate the efficacy of our system for tasks related to acute brain dysfunction including binary classification of brain acuity and multi-class classification (*i.e.*, coma, delirium, death, or normal), achieving a mean AUROC of 0.953 on our Longformer implementation. The deployment of our system for real-time prediction of ADB in ICUs can reduce the number of incidents caused by ABD. Moreover, the real-time system has the potential to reduce costs, duration of patients' stays in the ICU, and mortality among those afflicted.

## 1    Introduction

Assessing brain acuity status in the intensive care unit (ICU) to determine if a patient has normal brain activity or suffering acute brain dysfunction (ABD) (*i.e.*, coma and delirium, which can lead to death) is of great importance given this condition affects the majority (up to 64%) of critically ill patients at some point in their ICU stay [1-4] and is associated with higher mortality risk [5], longer hospital stays [6], and long-term cognitive impairment [7]. Current approaches for ABD diagnosis are limited to methods that assess current brain status or an overall risk assessment, which does not allow for early and dynamic diagnosis in the ICU. Such approaches include the use of assessment scores such as Glasgow Coma Scale (GCS) and Confusion Assessment Method (CAM), neuroimaging such as magnetic resonance imaging (MRI), and biomarkers such as S100 calcium-binding protein B (S100B) and neuron-specific enolase (NSE) [8]. These approaches are ineffective given they are based on observations by clinicians which are infrequent. A patient's condition in the ICU can change at any time, so it is important to capture the dynamic behavior of the patient to better assess the risk of ABD in the next hours and plan early interventions.

Given the wealth of data available in modern Electronic Health Record (EHR) systems, including vital signs, medications, laboratory results, assessment scores, and demographic data, there is an opportunity for use of data-driven approaches to dynamically predict ABD in the ICU. Deep learning has shown to have great predictive accuracy capacity in dynamic ICU outcome predictions, with models such as recurrent neural networks (RNNs) [9] and transformer models [10] being used for a variety of outcomes, including organ failure, Sequential Organ Failure Assessment (SOFA) scores, readmissions, and mortality prediction. These models can handle the large amounts of data present in EHR systems as well as patient data collected at admission and from the patient's medical history. This is especially true for patients in the ICU, as typical stays last about a week, with many lasting 2-3 weeks. This creates large sequences of data which deep learning models can process quickly. Therefore, these models have the potential to dynamically predict ABD in the ICU making use of the available data.

Despite the importance and prevalence of ABD in the ICU, there are very few studies on early prediction of this condition for earlier intervention. Most of the existing studies focus on predicting binary outcome for delirium (*i.e.*, delirium or non-delirium). In particular, there are several studies focusing on predicting the development of delirium at some point during the patient's stay in the ICU [11-14], postoperative delirium prediction [15], and predicting delirium dynamically in the ICU [16]. To the best of our knowledge, only two studies developed prediction models with multiple outcomes. The first study focused on predicting next-day outcome: either coma, delirium, death in ICU, discharge, or normal brain function [3]. The second study used the same model as the first one to predict transition from ABD state to ABD-free state [4].

## 1.1 Contributions

In this work, we provide multiple models for multi-class classification of brain acuity (*i.e.*, coma, delirium, death, or normal) dynamically to predict ABD in the ICU. Such models include baseline models such as Random Forest (RF), Extreme Gradient Boosting (XGB), CatBoost [17], and Gated Recurrent Unit (GRU), as well as Transformer models such as Longformer [18], Transformer-XL [19], Open Pre-trained Transformer (OPT) [20], Gated Transformer Network (GTN) [21], and Self-supervised Transformer for Time Series (STraTS) [22]. We provide robust models that outperform current models in the literature for ABD prediction in terms of sensitivity, specificity, and Area Under the Receiving Operator Curve (AUROC), and can make such predictions in real-time. To the best of our knowledge, this work uses the largest dataset for ABD prediction and most complex models compared to the state-of-the-art models. The models predict ABD in the next 12 hours making use of temporal data from EHRs pertaining to latest 12 hours of ICU stay, and static data from patient's information at admission and patient's medical history. Additionally, the models developed are tested for binary dynamic outcome for delirium in the ICU for comparison with current methods for delirium prediction.

## 2 Methods

## 2.1 Data Collection

The data used in this study was collected from patients in the ICU at the University of Florida Shands Hospital, from 2012 to 2019, with a total of 156,699 patients. The data was divided into two datasets: delirium dataset and brain acuity dataset.

The delirium dataset was composed of patients who were assessed using the Confusion Assessment Method (CAM) for delirium during their stay in the ICU. This dataset consists of binary labels for delirium, based on positive and negative CAM scores. For inclusion in the dataset, patients must have been in the ICU for at least 1 hour and contain at least 1 positive or negative CAM score during their ICU stay, which resulted in 21,321 patients being included in the dataset. If there are multiple encounters within a 24-hour period, they are merged into a single encounter. EHR data collected during each stay is as follows: laboratory test results (including test type and value), medications (medication type, dose value, dose unit), mobility score, American Society of Anesthesiologists (ASA) physical status classification, Braden scale, Glasgow Coma Scale (GCS), Modified early warning score (MEWS), Morse fall scale (MFS), Pain score, Richmond agitation-sedation scale (RASS), Sequential organ failure assessment (SOFA), blood pressure, heart rate, oxygen, respiratory rate, and temperature. This data along with demographic and medical history information from the patient is summarized in Table 1 in Section 2.3. The brain acuity dataset was composed of patients who were assessed using CAM, RASS, and GCS scores. Each nurse over a 12-hour shift assesses the patient's condition to obtain the mentioned scores and determine their brain acuity phenotype (*i.e.*, coma, delirium, normal, or dead). This dataset has significant overlap with the delirium dataset, with 21,107 patients appearing in both datasets. The CAM score is either positive (delirium) or negative (normal). The RASS score is a numerical number in the range [-5, 4], with 4 being "Combative" (physically violent), 0 being normal, -3 being moderate sedation, and –5 being unarousable. The GCS score is a numerical score, using a summation of points on a variety of tests of the patient's response to stimuli. The maximum GCS score is 15, which attributes to the normal state, less than or equal 8 being comatose, and less than 3 being unresponsive. Any missing scores are imputed by carrying forward the previous measurement as long as it occurred within 12 hours of the missing value. If no score is available, the value is marked as missing. If all three scores are missing for a 12-hour shift, then the shift is excluded. The logic for calculating brain acuity phenotype is shown in Fig. 1. This logic was created by us based on CAM, RASS, and GCS scores [23]. Normal brain status has a RASS score greater than –3 and a negative CAM score. If RASS is less than –3, or GCS less than or equal to 8 when RASS score is missing, then the patient is determined to be in a coma state. For RASS scores equaling –3, the GCS score is used as the tie breaker between delirium and coma.

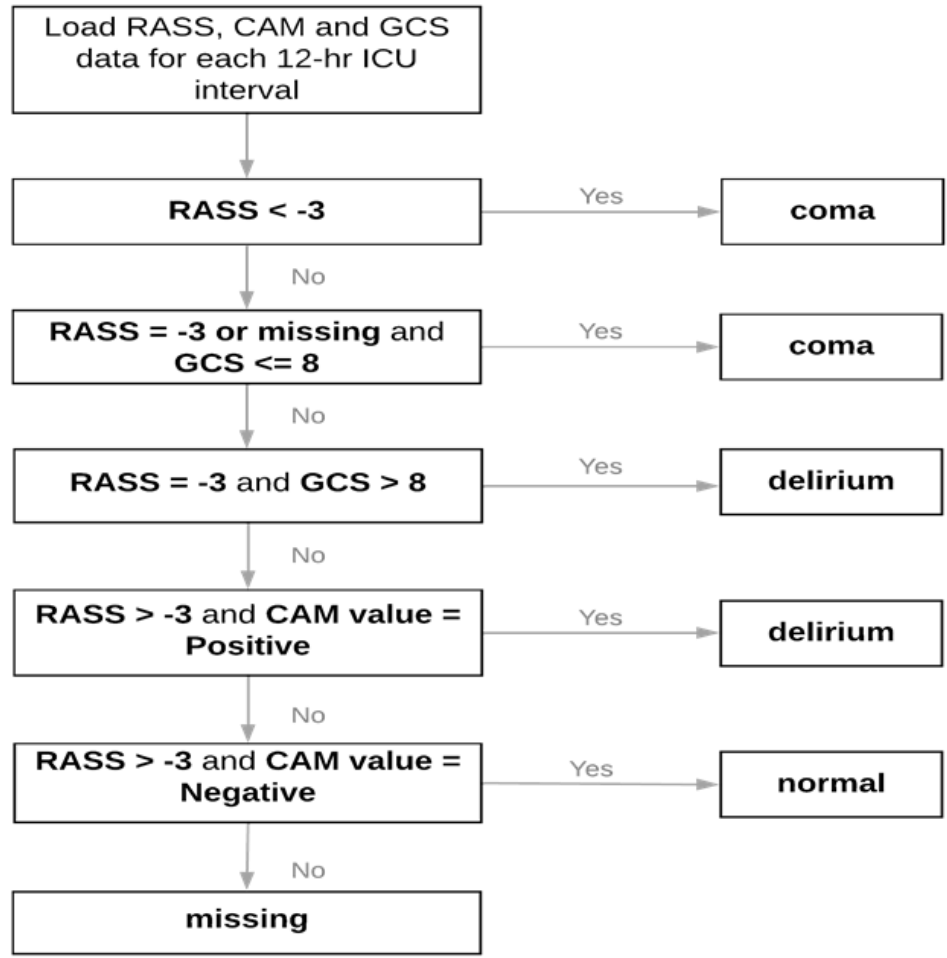

Fig. 1: Brain status phenotype logic diagram [23].

In terms of statistics for the data extracted in the delirium dataset, the median duration of an ICU stay was about 8 days. The delirium dataset had a median 4 CAM scores recorded per stay for each patient. This dataset had a delirium incidence of 8%. The brain acuity dataset also had a median ICU stay duration of 8 days, with about 80 brain acuity phenotypes recorded during each ICU stay. This dataset contains 6% delirium, 3% mortality, and 9% coma.

## 2.2   Data Extraction and Processing

To process the large number of features of the EHRs of each patient, a specialized data processing pipeline was developed to extract, filter, normalize, and combine all of the features into a model-readable input. For the Transformer models used in this study, we have modified the data processing pipeline according to each model's requirement with respect to fixed and variable sequence lengths, maximum number of tokens, and whether the model can process sequences of temporal data.

For Longformer, Transformer-XL, and OPT the processing was as follows. Each feature has the top and bottom 1% of outliers removed. Medications that occur less than 1% in the dataset are dropped, with the same occurring for labs. Medication names and laboratory test names were standardized to remove duplicate names. Then sequences of lengths greater than 12000 are clipped, which is the top 1% of the sequence lengths. Each feature is then imputed using the median for numerical features and most frequent occurrence for categorical features, normalized to a standard distribution. Once the dataset is combined, positional tokens are added to the temporal features and packaged using the Hierarchical Data Format 5 (hdf5) format for model training.

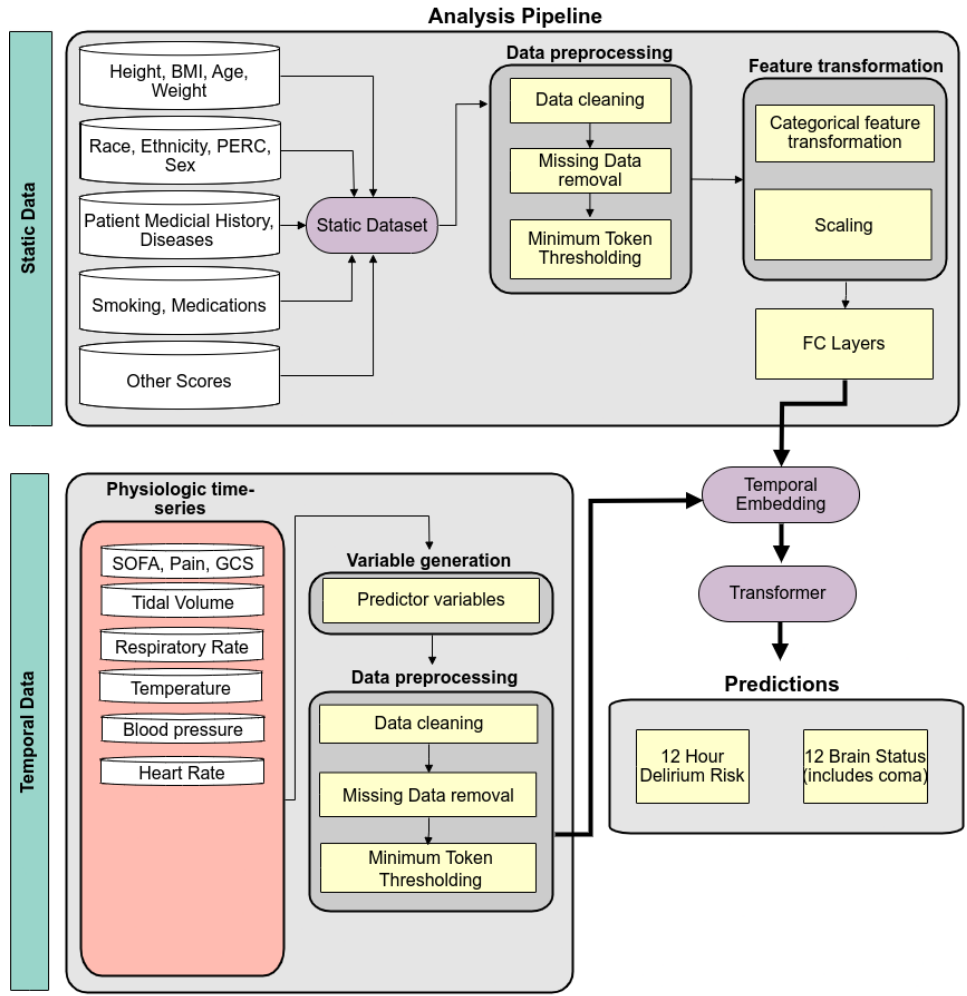

Fig. 2: Data Analysis Pipeline showing the temporal and static data processing.

For data extraction of the brain acuity phenotype labels, hospital encounter IDs with brain acuity recorded in 7 AM to 7 PM and 7 PM to 7 AM shifts were extracted from the University of Florida Hospital database. A first filter was applied based on shifts that started within an intensive care unit (ICU) stay (*i.e.*, after admission and before discharge). Then, a second filter was applied to drop any ICU stay that lasted less than 12 hours and drop any shifts that started less than 12 hours after admission such that the models could have the latest 12 hours of ICU observations available to make a prediction. The final dataset which consisted of 36,083 patients with 49,936 ICU stays and 496,437 shifts with brain acuity recorded. The dataset was split into train, validation and test set based on patient IDs.

For baseline models and Gated Transformer Network (GTN) [21] model, the data was tabularized such that each column represented one feature. Missing values were imputed by first performing linear interpolation on individual ICU stays and propagating backward and forward values where the conditions for interpolation were not satisfied. Remaining missing values were then imputed using mean method for any non-medication variables and using zero for medication values. All values were then scaled to a [0, 1] range.

For the Self-Supervised Transformer for Time Series (STraTS) [22] model, the dataset was imputed using most frequent method for categorical values and median method for numeric values. Then, standard scaling was used to standardize all values. Variables were assigned a variable code for embedding.

A schematic of the data extraction and processing pipeline can be seen in Fig. 3.

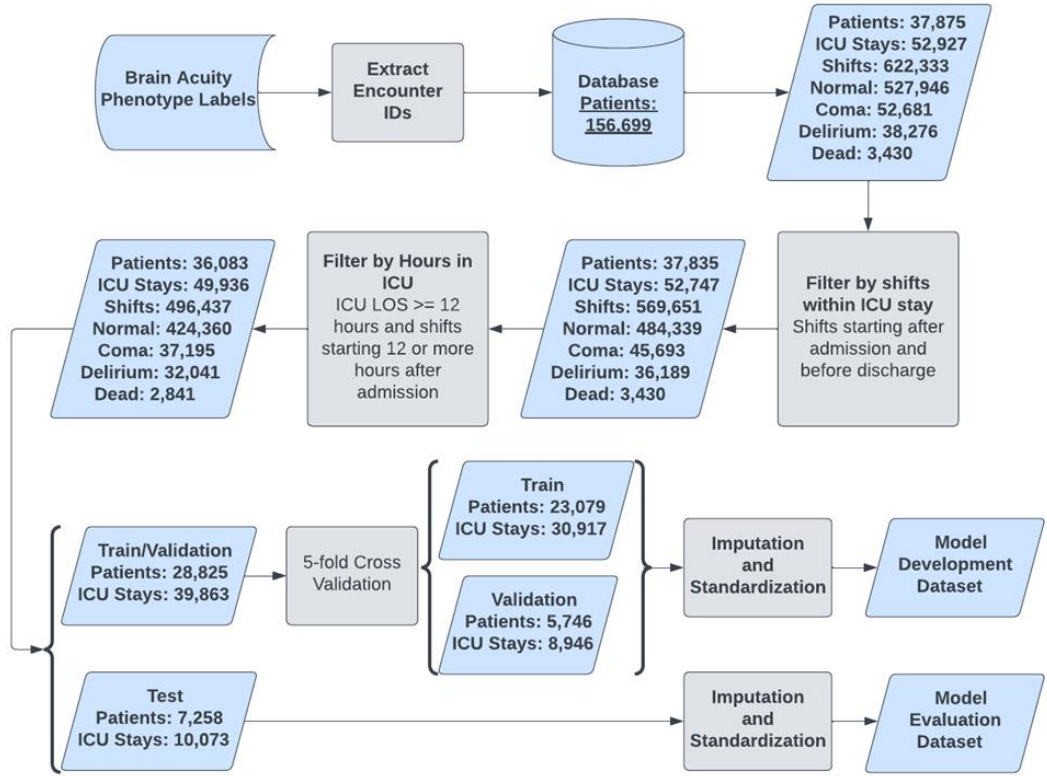

Fig. 3: Data extraction and processing pipeline. Stored data for brain acuity phenotype labels is used to extract hospital encounter IDs to query the database and get the initial data. The first filter drops any shift which did not occur during an ICU stay and the second filter drops any ICU stay with length of stay (LOS) less than 12 hours and shifts which started less than 12 hours after ICU admission. The dataset is then split into train/validation and test sets based on patient IDs. The train/validation set is further split for 5-fold cross validation. Finally, imputation and standardization are applied to all sets to obtain the model development and model evaluation datasets.

## 2.3  Feature extraction

The features used for model training included temporal and static data. The temporal data was extracted from the EHR system as described in Section 2.2 and consisted of 6 vital signs measurements, a total of 166 unique medications and 101 laboratory results after filtering medications and lab tests present in less than 5% of ICU stays, and 30 ICU assessment scores which included all subcomponents of each method. The static data consisted of 593 patient characteristics including demographics, admission information, comorbidities, admission type, neighborhood characteristics, and patient history from the previous year which included medications and laboratory tests. This resulted in a total of 896 features. A summary of these features can be seen in Table 1.

| Category | Variable | Type |
|---|---|---|
| **Patient demographics** | Age, Sex, Ethnicity, Race, Language, Marital status, Smoking status, Insurance provider | Static |
| **Patient residential information** | Total population, Distance from hospital, Rural/Urban, Median income, Proportion black, Proportion Hispanic, Percent below poverty line | Static |
| **Patient admission information** | Height, Weight, Body mass index, 17 comorbidities present at Admission, Charlson comorbidity index, Presence of chronic kidney disease, Admission type | Static |
| **Patient medication history** | ACE inhibitors, Aminoglycosides, Antiemetics, Aspirin, Beta blockers, Bicarbonates, Corticosteroids, Diuretics, NSAIDS, Vasopressors/Inotropes, Statins, Vancomycin, Nephrotoxic drugs | Static |
| **Patient laboratory test results history** | Serum hemoglobin, Urine hemoglobin, Serum glucose, Urine glucose, Urine red blood cells, Urine protein, Serum urea nitrogen, Serum creatinine, Serum calcium, Serum sodium, Serum potassium, Serum chloride, Serum carbon dioxide, White blood cells, Mean corpuscular volume, Mean corpuscular hemoglobin, Hemoglobin concentration, Red blood cell distribution, Platelets, Mean platelet volume, Serum anion gap, Blood pH, Serum oxygen, Bicarbonate, Base deficit, Oxygen saturation, Band count, Bilirubin, C-reactive protein, Erythrocyte sedimentation rate, Lactate, | Static |
| | Troponin T/I, Albumin, Alaninen, Asparaten | |
| **ICU vital signs** | Systolic blood pressure, Diastolic blood pressure, Heart rate, Respiratory rate, Oxygen saturation | Temporal |
| **ICU Assessment Scores** | ASA physical status classification, Braden scale, Modified early warning score (MEWS), Morse fall scale (MFS), Pain score, Richmond agitation-sedation scale (RASS), Sequential organ failure assessment (SOFA), Glasgow Coma Scale (GCS), Mobility assistance level | Temporal |
| **ICU laboratory tests** | 101 distinct laboratory tests | Temporal |
| **ICU medications** | 166 distinct medications | Temporal |

### 2.4 Baseline Models

To compare the performance of the Transformer models with commonly used classification models, four baseline models were developed: Random Forest (RF), XGBoost (XGB), CatBoost, and Gated Recurrent Unit (GRU). For RF, CatBoost, and XGB models, the average for each temporal feature in the latest 12 hours was taken and used as input along the static data. For the GRU model, the temporal data was fed through a single GRU layer, from which the output was concatenated with the static data and fed through two fully connected layers. The same processing was used for both delirium and brain status outcomes.

### 2.5 Transformer Models

We developed and tested several Transformer models on the delirium and the brain activity datasets. In what follows, we provide a brief description of each model. For a more detailed description of each model, we refer the reader to the reader to the respective references.

Gated Transformer Network (GTN) [21], uses a combination of two embeddings. The first embedding is created step-wise, meaning the embedding is created along the timesteps dimension. The second embedding is created channel-wise, meaning the embedding is created along the features dimension. This model requires data to be tabularized the same way as with baseline models. Therefore, the GTN model provides a more direct comparison with baseline models. Another transformer model included in this study is the Self-supervised Transformer for Time Series (STraTS) [22]. The advantage of the STraTS model relies on the usage of triplet embeddings, which removes the need of tabularizing the data and performing excessive imputation. We adapted the architecture of the STraTS model for the delirium and brain acuity prediction tasks. The model splits patient data for a particular 12-hour shift in the ICU into static and temporal data. The static data goes through a feed forward network (FFN) to create the static embedding. The temporal data contains $n$ number of observations, where each observation has a timestamp, variable code, and value. Each of the three are embedded: timestamp and value are embedded using a one-to-many network, and variable code is embedded using an embedding layer. The three embeddings are fused through addition to be passed through $M$ transformer layers to create contextual embeddings. These contextual embeddings are fused to create a temporal embedding. The temporal and static embedding are concatenated and passed through a dense layer with four output neurons to predict probabilities of each class (*i.e.*, dead, coma, delirium, and normal). Predicted brain acuity is determined by the class with the highest probability. The overall architecture of the model can be seen in Fig. 4.

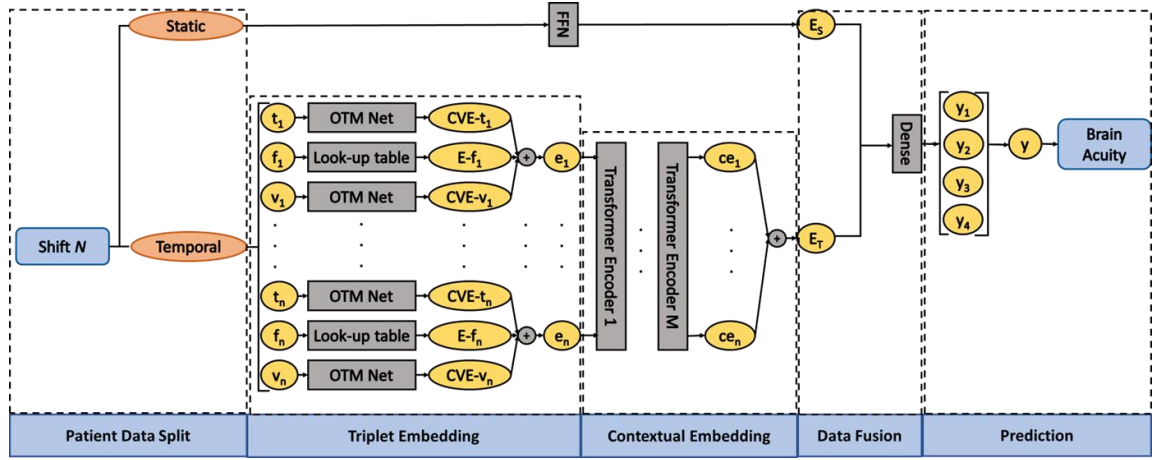

Fig. 4: Schematic for one of the models for brain acuity prediction. For each shift $N$, temporal and static data were split. Temporal data consisting of $n$ observations was embedded. The embedding consisted of taking timestamp ($t$), variable/feature code ($f$), and value ($v$) at each observation and creating a continuous value embedding for timestamp (*CVE-t*) and value (*CVE-v*) using a one-to-many network (*OTM Net*) and an embedding for feature (*E-f*) using an embedding layer (*i.e.*, look-up table). These three embeddings are fused into a single embedding ($e$) for each observation, and the embeddings were then passed through $M$ transformer layers to create contextual embeddings (*ce*). All contextual embeddings are fused to create a temporal embedding ($E_T$) which is then combined with a static embedding ($E_S$) created by passing the static data through a feed forward network (*FFN*). The fused embedding is then passed through a dense layer to predict probabilities for the four brain acuity labels ($y_1$, $y_2$, $y_3$, $y_4$) and a final prediction for brain acuity is made by picking the class with highest probability ($y$).

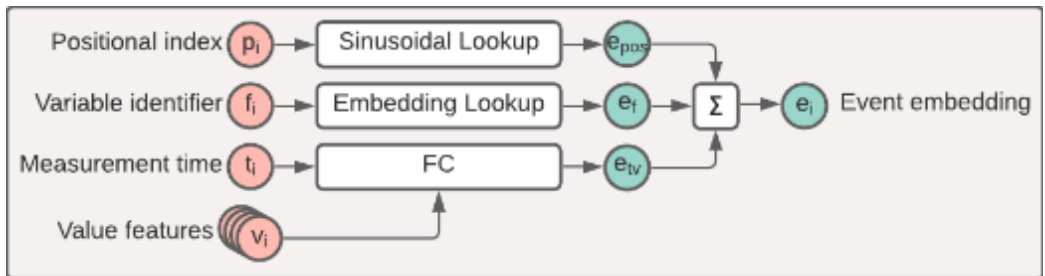

Fig. 5: Embedding structure for temporal data of Transformer models.

Other Transformer models considered in this study are Longformer, Open Pre-trained Transformer (OPT), and Transformer-XL. These models were trained from scratch using their default hyperparameters. The dataset was processed as described in section 2.2 and passed through our novel embedding layers (Fig. 4 and Fig. 5) before input into each of these models. These models were chosen since they all have different advantages when it comes to our dataset. The ability of the Longformer model to handle large sequences with many features efficiently makes it well-suited for our dataset. This advantage is due to the design of its self-attention mechanism, which combines both a local and global attention window. The OPT model, based off of OpenAI's GPT2, which offers similar performance to OpenAI's GPT3 with reduced costs to train, is a large model built originally for language modeling but can be adapted for a variety tasks. The advantage of this model is its large parameter space allowing to capture complex relationships among the data. The Transformer-XL model removes the dependency for fixed length input sequences, allowing for modeling of longer-range dependencies among the input data, using a segment-level recurrence scheme.

## 3    Results

All models are trained on the delirium dataset for binary outcome prediction, and brain acuity dataset for ABD prediction. For ABD prediction, the overall performance of the models is computed using Area Under the Receiving Operator Curve (AUROC), and individual performance for each class is also computed. Metrics used for performance assessment include AUROC, Area Under the Precision Recall Curve (AUPRC), Sensitivity, Specificity, Positive Predictive Value (PPV), and Negative Predictive Value (NPV). All metrics are computed by taking the average across 5-fold cross validation with 10-iteration bootstrap for each fold, and the 95% Confidence Interval (CI) are calculated across all repetitions.

The results for delirium prediction showed better performance of the transformer models over baseline models in terms of Area Under the Receiving Operator Curve (AUROC) as seen in Table 2. However, the STraTS model had a significantly lower performance than baseline models which could be due to the different processing and embedding that it uses compared to baseline and other Transformer models. The best performing model on all metrics was the Longformer (0.870 AUROC), while baseline models were not far away in performance (AUROC ranging from 0.830-0.850).

Table 2. Results from 5-fold cross validation with 10-iteration bootstrap for delirium prediction.

| Model | AUROC (95% CI) | AUPRC (95% CI) | Sensitivity (95% CI) | Specificity (95% CI) | PPV (95% CI) | NPV (95% CI) |
|---|---|---|---|---|---|---|
| **RF** | 0.834 (0.830-0.847) | 0.268 (0.251-0.307) | 0.858 (0.835-0.884) | 0.679 (0.651-0.708) | 0.179 (0.161-0.214) | 0.983 (0.982-0.986) |
| **XGB** | 0.830 (0.824-0.842) | 0.263 (0.248-0.296) | 0.857 (0.824-0.884) | 0.672 (0.644-0.702) | 0.176 (0.160-0.209) | 0.983 (0.981-0.986) |
| **CatBoost** | 0.850 (0.841-0.855) | 0.269 (0.252-0.295) | 0.858 (0.826-0.885) | 0.683 (0.654-0.705) | 0.181 (0.161-0.210) | 0.983 (0.981-0.986) |
| **GRU** | 0.841 (0.826-0.860) | 0.289 (0.250-0.345) | 0.858 (0.838-0.896) | 0.690 (0.658-0.723) | 0.184 (0.165-0.223) | 0.983 (0.981-0.986) |
| **GTN** | 0.832 (0.818-0.854) | 0.267 (0.233-0.326) | 0.864 (0.833-0.883) | 0.672 (0.637-0.714) | 0.177 (0.154-0.219) | 0.983 (0.982-0.986) |
| **STraTS** | 0.806 (0.797-0.814) | 0.205 (0.192-0.216) | 0.837 (0.817-0.856) | 0.647 (0.614-0.672) | 0.143 (0.134-0.163) | 0.982 (0.982-0.984) |
| **Long-former** | **0.870 (0.840-0.890)** | **0.343 (0.322-0.365)** | **0.904 (0.880 - 0.920)** | **0.725 (0.713 - 0.733)** | **0.203 (0.198 - 0.207)** | **0.984 (0.983 - 0.985)** |
| **OPT** | 0.864 (0.835-0.885) | 0.332 (0.312-0.353) | 0.895 (0.878 - 0.913) | 0.712 (0.702 - 0.721) | 0.199 (0.195 - 2.04) | **0.984 (0.983 - 0.985)** |

| Transfo-XL | 0.859 (0.835-0.875) | 0.320 (0.290-0.342) | 0.860 (0.842 - 0.879) | 0.704 (0.698 - 0.709) | 0.196 (0.193 - 0.198) | **0.984 (0.983 - 0.985)** |
|---|---|---|---|---|---|---|

For brain acuity prediction, overall performance was measured by AUROC. As seen in Table 3, Longformer had the highest mean AUROC (0.953) as well as individual AUROC for each class (0.983 for coma, 0.887 for delirium, and 0.989 for mortality). Metrics for individual classes were also computed. For coma prediction, Longformer had highest performance for most metrics except for specificity (GTN and OPT) and negative predictive value (NPV) (STraTS) as seen in Table 4. For delirium prediction, Longformer also had highest performance for most metrics except for sensitivity (GTN) and Area Under the Precision Recall Curve (AUPRC) (OPT) as seen in Table 5. For mortality prediction, Longformer had the highest performance for most metrics except for sensitivity (XGB) and AUPRC (CatBoost) as seen in Table 6.

Table 3. Brain acuity prediction results from 5-fold cross validation with 10-iteration bootstrap expressed as area under the receiving operator characteristic curve (AUROC).

| Model | Mean (95% CI) | Coma (95% CI) | Delirium (95% CI) | Mortality (95% CI) |
|---|---|---|---|---|
| **RF** | 0.897 (0.887-0.909) | 0.953 (0.948-0.959) | 0.783 (0.769-0.804) | 0.956 (0.945-0.964) |
| **XGB** | 0.906 (0.899-0.917) | 0.948 (0.941-0.955) | 0.802 (0.792-0.821) | 0.970 (0.965-0.975) |
| **CatBoost** | 0.909 (0.905 - 0.915) | 0.945 (0.939 - 0.950) | 0.810 (0.804 - 0.815) | 0.971 (0.967 - 0.975) |
| **GRU** | 0.897 (0.881-0.913) | 0.955 (0.945-0.962) | 0.788 (0.758-0.819) | 0.950 (0.940-0.959) |
| **GTN** | 0.885 (0.858-0.907) | 0.933 (0.901-0.956) | 0.784 (0.755-0.815) | 0.939 (0.919-0.951) |
| **STraTS** | 0.842 (0.789-0.875) | 0.902 (0.850-0.931) | 0.690 (0.594-0.750) | 0.936 (0.925-0.945) |
| **Long-former** | **0.953 (0.945 - 0.966)** | **0.983 (0.980 - 0.987)** | **0.887 (0.870 - 0.918)** | **0.989 (0.985 - 0.993)** |
| **OPT** | 0.945 (0.930 - 0.954) | 0.980 (0.975 - 0.983) | 0.868 (0.831 - 0.889) | 0.987 (0.984 - 0.990) |
| **Transfo-XL** | 0.934 (0.925 - 0.946) | 0.975 (0.970 - 0.979) | 0.857 (0.840 - 0.886) | 0.970 (0.965 - 0.973) |

Table 4. Coma class prediction results from 5-fold cross validation with 10-iteration bootstrap.

| Model | AUROC (95% CI) | AUPRC (95% CI) | Sensitivity (95% CI) | Specificity (95% CI) | PPV (95% CI) | NPV (95% CI) |
|---|---|---|---|---|---|---|
| **RF** | 0.953 (0.948-0.959) | 0.698 (0.669-0.756) | 0.660 (0.620-0.689) | 0.954 (0.944-0.964) | 0.591 (0.533-0.643) | 0.965 (0.961-0.972) |
| **XGB** | 0.948 | 0.702 (0.673-0.751) | 0.698 (0.666-0.723) | 0.959 (0.954-0.966) | 0.635 (0.604-0.694) | 0.969 (0.965-0.974) |

| | | | | | |
|---|---|---|---|---|---|
| | (0.941-0.955) | | | | | |
| **CatBoost** | 0.945 (0.939 - 0.950) | 0.696 (0.667 - 0.732) | 0.692 (0.663 - 0.715) | 0.954 (0.945 - 0.961) | 0.632 (0.615 - 0.649) | 0.966 (0.963 - 0.969) |
| **GRU** | 0.955 (0.945-0.962) | 0.721 (0.660-0.770) | 0.678 (0.600-0.750) | 0.960 (0.931-0.974) | 0.635 (0.501-0.715) | 0.967 (0.956-0.976) |
| **GTN** | 0.933 (0.901-0.956) | 0.672 (0.546-0.776) | 0.632 (0.457-0.745) | **0.964 (0.936-0.983)** | 0.653 (0.527-0.772) | 0.963 (0.944-0.976) |
| **STraTS** | 0.902 (0.850-0.931) | 0.509 (0.455-0.575) | 0.643 (0.592-0.716) | 0.940 (0.914-0.968) | 0.485 (0.388-0.611) | **0.970 (0.963-0.977)** |
| **Long-former** | **0.983 (0.980 - 0.987)** | **0.773 (0.703 - 0.855)** | **0.740 (0.611 - 0.774)** | 0.957 (0.929 - 0.971) | **0.678 (0.593 - 0.732)** | 0.968 (0.960 - 0.976) |
| **OPT** | 0.980 (0.975 - 0.983) | 0.765 (0.694 - 0.832) | 0.737 (0.643 - 0.766) | **0.964 (0.939 - 0.976)** | 0.661 (0.601 - 0.727) | 0.964 (0.959 - 0.968) |
| **Transfo-XL** | 0.975 (0.970 - 0.979) | 0.764 (0.694 - 0.811) | 0.733 (0.655 - 0.755) | 0.950 (0.939 - 0.960) | 0.659 (0.605 - 0.722) | 0.964 (0.959 - 0.968) |

Table 5. Delirium class prediction results from 5-fold cross validation with 10-iteration bootstrap.

| Model | AUROC (95% CI) | AUPRC (95% CI) | Sensitivity (95% CI) | Specificity (95% CI) | PPV (95% CI) | NPV (95% CI) |
|---|---|---|---|---|---|---|
| **RF** | 0.783 (0.769-0.804) | 0.222 (0.185-0.264) | 0.652 (0.567-0.716) | 0.764 (0.730-0.795) | 0.184 (0.166-0.223) | 0.964 (0.961-0.969) |
| **XGB** | 0.802 (0.792-0.821) | 0.238 (0.204-0.291) | 0.678 (0.627-0.724) | 0.768 (0.746-0.793) | 0.192 (0.175-0.230) | 0.967 (0.963-0.970) |
| **CatBoost** | 0.810 (0.804 - 0.815) | 0.250 (0.217 - 0.310) | 0.687 (0.634 - 0.723) | 0.774 (0.744 - 0.798) | 0.193 (0.185 - 0.229) | 0.976 (0.969 - 0.984) |
| **GRU** | 0.788 (0.758-0.819) | 0.210 (0.163-0.267) | 0.694 (0.570-0.790) | 0.738 (0.693-0.796) | 0.178 (0.153-0.222) | 0.967 (0.959-0.975) |
| **GTN** | 0.784 (0.755-0.815) | 0.197 (0.161-0.249) | **0.704 (0.551-0.815)** | 0.726 (0.642-0.812) | 0.174 (0.154-0.214) | 0.968 (0.956-0.979) |
| **STraTS** | 0.690 (0.594-0.750) | 0.128 (0.087-0.158) | 0.543 (0.339-0.729) | 0.734 (0.672-0.831) | 0.127 (0.087-0.150) | 0.959 (0.948-0.972) |

| **Long-former** | **0.887 (0.870 - 0.918)** | 0.262 (0.202 - 0.313) | 0.699 (0.645 - 0.853) | **0.832 (0.788 - 0.875)** | **0.213 (0.175 - 0.222)** | **0.988 (0.985 - 0.993)** |
|---|---|---|---|---|---|---|
| **OPT** | 0.868 (0.831 - 0.889) | **0.265 (0.202 - 0.321)** | 0.684 (0.593 - 0.756) | 0.829 (0.799 - 0.854) | 0.199 (0.166 - 0.235) | 0.981 (0.977 - 0.984) |
| **Transfo-XL** | 0.857 (0.840 - 0.886) | 0.255 (0.195 - 0.294) | 0.689 (0.544 - 0.798) | 0.801 (0.759 - 0.832) | 0.179 (0.159 - 0.212) | 0.978 (0.974 - 0.984) |

Table 6. Mortality class prediction results from 5-fold cross validation with 10-iteration bootstrap.

| **Model** | **AUROC (95% CI)** | **AUPRC (95% CI)** | **Sensitivity (95% CI)** | **Specificity (95% CI)** | **PPV (95% CI)** | **NPV (95% CI)** |
|---|---|---|---|---|---|---|
| **RF** | 0.956 (0.945-0.964) | 0.225 (0.195-0.260) | 0.775 (0.713-0.834) | 0.943 (0.927-0.960) | 0.070 (0.053-0.082) | 0.998 (0.999-0.999) |
| **XGB** | 0.970 (0.965-0.975) | 0.302 (0.262-0.341) | **0.813 (0.758-0.853)** | 0.946 (0.932-0.963) | 0.076 (0.064-0.094) | 0.998 (0.999-0.999) |
| **CatBoost** | 0.971 (0.967 - 0.975) | **0.305 (0.255 - 0.354)** | 0.811 (0.765 - 0.849) | 0.946 (0.935 - 0.967) | 0.080 (0.060 - 0.96) | 0.998 (0.999 - 0.999) |
| **GRU** | 0.950 (0.940-0.959) | 0.178 (0.143-0.218) | 0.673 (0.505-0.795) | 0.946 (0.893-0.975) | 0.069 (0.041-0.102) | 0.998 (0.997-0.999) |
| **GTN** | 0.939 (0.919-0.951) | 0.171 (0.130-0.214) | 0.697 (0.505-0.848) | 0.936 (0.869-0.980) | 0.065 (0.033-0.120) | 0.998 (0.997-0.999) |
| **STraTS** | 0.936 (0.925-0.945) | 0.140 (0.116-0.172) | 0.663 (0.529-0.814) | 0.934 (0.872-0.971) | 0.069 (0.037-0.095) | 0.997 (0.997-0.999) |
| **Long-former** | **0.989 (0.985 - 0.993)** | 0.248 (0.199 - 0.345) | 0.779 (0.569 - 0.880) | **0.954 (0.895 - 0.985)** | **0.091 (0.081 - 0.103)** | **0.999 (0.998-0.999)** |
| **OPT** | 0.987 (0.984 - 0.990) | 0.276 (0.213 - 0.330) | 0.770 (0.689 - 0.857) | 0.946 (0.912 - 0.989) | 0.088 (0.064 - 0.112) | 0.998 (0.998-0.999) |
| **Transfo-XL** | 0.970 (0.965 - 0.973) | 0.221 (0.165 - 0.289) | 0.748 (0.649 - 0.842) | 0.948 (0.892 - 0.982) | 0.088 (0.078 - 0.098) | 0.996 (0.995-0.997) |

To better visualize the classes that models find the most difficult to distinguish from one another, the confusion matrix for recall (*i.e.,* sensitivity) for the XGB model in the brain acuity dataset is depicted in Fig. 6.

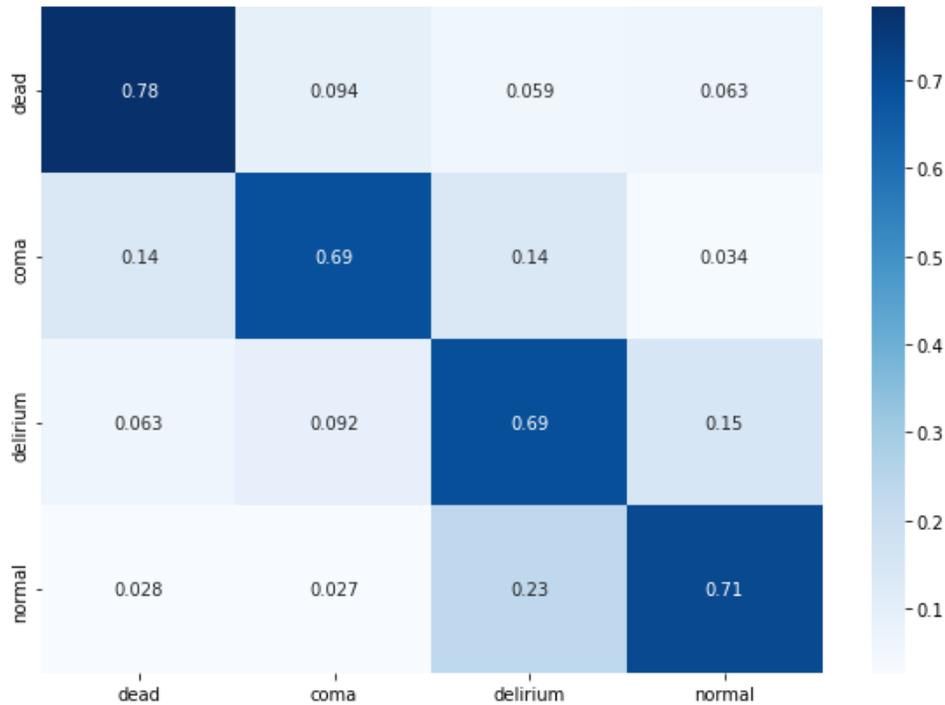

Fig. 6: Recall confusion matrix for brain acuity outcome prediction for XGB model

## 4    Discussion

Our results from this experiment are consistent with our hypothesis; that Transformer models can process patient's EHR data to accurately predict the onset of ABD. Transformer models improve over baseline approaches such as CatBoost, XGB, and GRU models for ABD prediction. Furthermore, as these models use only the EHR data, the data processing pipeline and the models can be adopted by other healthcare providers.

In comparison to current ABD early prediction approaches in the literature, our models achieve greater sensitivity, specificity, and AUROC [3, 4]. Our models have the advantage of using a larger dataset (36,083 patients vs 810 patients [3] and 1,026 patients [4]) and more complex architectures in terms of the Transformer models. However, the prediction on the delirium dataset only seemed to offer little improvement over baseline approaches we have tested and did not improve performance compared to recent dynamic ICU delirium models in terms of AUROC (Longformer achieved 0.870 compared to up to 0.909 on recent study [16]). This may be due to the smaller dataset size (21,321 patients in our dataset compared to 38,436 patients on recent study [16]) and lower delirium incidence (8% in our dataset compared to 54.5% in recent study [16]). The brain acuity dataset showed better performance for delirium prediction which can be due to the larger size of the dataset and higher delirium incidence. Additional data from other sources, such as the public MIMIC-IV and eICU datasets, can further improve the performance of Transformer models over the baseline models. The use of data collected from different environments has the potential to improve the generalizability of the models to other healthcare settings. Looking at the baseline models performance on the brain acuity dataset showed that the models have the highest error in distinguishing coma and delirium outcomes, as well as normal and delirium as depicted in the recall confusion matrix in Fig. 6. Delirium affects people at different levels, with some experiencing milder symptoms and others experiencing more extreme symptoms. As more extreme delirium cases may be similar to the coma outcomes and milder delirium cases is more similar to the normal ones, their distinguishment is more difficult. Hence, the higher error rate of the models for these cases in comparison to others may be due to the more difficult nature of such classification. With adaptation to a real-time data pipeline, these models can be deployed to calculate ABD onset in practice with enough confidence to help clinicians prepare and treat ABD for those afflicted in the ICU.

## 5    Conclusion

Early detection of Acute brain dysfunctions (ABD) in ICUs is critical for timely intervention. The absence of such early detection could lead to severe outcomes, including death. Current systems to detect ABD require manual assessment and review from clinical staff, and can only detect ABD when onset is already occurring. An AI-enabled automated ABD monitoring and prediction system can prevent negative outcomes, reduce costs, and lead to shorter ICU stays. In this work, we adopt and test several machine learning models to predict an onset of ABD 12 hours in advance using the EHR data from ICU patients at the UF Shands Hospital. Among these models, Transformer models exhibited the best performance on a variety of tasks related to ABD prediction. The real-time deployment of these models has the potential to provide early warnings to the clinical staff on ABD onset and help save time, reduce costs, and save lives. Future work will further improve and validate our approaches using publicly available datasets to ensure the robustness of the models and their generalizability to other hospital settings.

## 6    Conflict of Interest

The authors declare that the research was conducted in the absence of any commercial or financial relationships that could be construed as a potential conflict of interest.

## 7    Funding

A.B. and P.R. were supported by NIH/NINDS R01 NS120924, NIH/NIBIB R01 EB029699, NIH/NIGMS R01 GM110240, and NIH Bridge2AI OT2OD032701, as well as the NIH/NCATS Clinical and Translational Sciences Award to the University of Florida UL1 TR000064. A.B. was supported by NIH/NIDDK R01 DK121730. P.R. was supported by NIH/NIBIB R21 EB027344 and NSF CAREER 1750192.

## 8    Data Availability Statement

The data used in this study is not publicly available. It was collected and processed by the I-HEAL lab at UF from participating patients in UF Shands hospital.